\documentclass{tlp}
\usepackage{mathptmx}
\usepackage[disable]{todonotes}
\usepackage{multicol}
\usepackage{times}
\usepackage{helvet}
\usepackage{courier}
\usepackage{xspace}
\usepackage{latexsym}
\usepackage{enumitem}
\usepackage{verbatim}
\usepackage{amsmath}
\usepackage{amssymb}
\usepackage{color}
\usepackage{tikz}
\usetikzlibrary{shapes,arrows}
\usepackage{hyperref}

\newtheorem{exmp}{Example}[section]

\renewcommand{\paragraph}[1]{
	
	\medskip
	\noindent 
	{\bf #1}~}

\def\beq{\begin{equation}}
\def\eeq#1{\label{#1}\end{equation}}
\def\ba{\begin{array}}
	\def\ea{\end{array}}

\def\idlv{{\sc idlv}\xspace}

\def\lpopt{{\sc lpopt}\xspace}
\def\gringo{{\sc gringo}\xspace}
\def\clingo{{\sc clingo}\xspace}
\def\pyclingo{{\sc pyclingo}\xspace}

\def\dlv{{\sc dlv}\xspace}

\def\pr{{\sc projector}\xspace}
\def\prdpr{{\sc prd-projector}\xspace}
\def\prdprshort{{\sc prd-proj}\xspace}
\def\prshort{{\sc proj}\xspace}
\def\prd{{\sc predictor}\xspace}
\def\prdlpopt{{\sc prd-lpopt}\xspace}

\def\aspccg{{\sc aspccg}\xspace}

\def\enco{{\sc enc1}\xspace}
\def\encm{{\sc enc7}\xspace}
\def\encl{{\sc enc19}\xspace}

\definecolor{Gray}{gray}{0.75}

\def\ar{\leftarrow}

\def\beq{\begin{equation}}
\def\eeq#1{\label{#1}\end{equation}}
\def\ba{\begin{array}}
	\def\ea{\end{array}}
\def\<{\langle}
\def\>{\rangle}

\mathchardef\mhyphen="2D
\def\citeb#1{(\citeauthor{#1}, \citeyear{#1})}

\def\gringo{{\sc gringo}\xspace}

\def\dlv{{\sc dlv}}

\definecolor{Gray}{gray}{0.75}

\newcommand{\ignore}[1]{}

\def\vars{\mathit{vars}}
\def\kvars{\mathit{kvars}}
\def\args{\mathit{args}}

\def\ar{\leftarrow}

\def\beq{\begin{equation}}
\def\eeq#1{\label{#1}\end{equation}}

\begin{document}
\lefttitle{Daniel Bresnahan, Nicholas Hippen, Yuliya Lierler}

\title[Predictor: Grounding Size Estimator]{
System Predictor: Grounding Size Estimator for\\ Logic Programs under Answer Set Semantics}
%
%
\begin{authgrp}
\author{ \gn{Daniel Bresnahan}}
\affiliation{University of Nebraska Omaha}
\author{ \gn{Nicholas Hippen}}
\affiliation{University of Nebraska Omaha}
\author{ \gn{Yuliya Lierler}}
\affiliation{University of Nebraska Omaha}
\end{authgrp}

\jnlPage{\pageref{firstpage}}{\pageref{lastpage}}
\jnlDoiYr{2021}
\doival{10.1017/xxxxx}

\maketitle

%
%

%
\begin{abstract}
Answer set programming  is a declarative logic programming paradigm geared towards solving difficult combinatorial search problems. While different logic programs can encode the same problem, their performance may vary significantly. It is not always easy to identify which version of the program performs the best. We present the system \prd (and its algorithmic backend) for estimating the grounding size of programs, a metric that can influence a performance of a system processing a program. We evaluate the impact of \prd when used as a guide for rewritings produced by the answer set programming  rewriting tools \pr and \lpopt. The results demonstrate potential to this approach.
Under consideration in Theory and Practice of Logic Programming (TPLP).

\end{abstract}
\section{Introduction}

Answer set programming (ASP) \citeb{BrewkaET11} is a declarative (constraint) programming paradigm geared towards solving difficult combinatorial search problems. ASP programs model problem specifications/constraints as a set of logic rules. These logic rules define a problem instance to be solved. An ASP system is then used to compute  solutions (answer sets) to the program. 
Answer set programming has been successfully used 
in scientific and industrial applications. 
Examples include, but are not limited to a
decision support systems for 
space shuttle flight controllers~\citeb{BalducciniGN06},
 team building and scheduling~\citeb{RiccaGAMLIL12}, and healthcare realm~\citeb{dod21}. 

\begin{figure}[t]
	\centering
	\includegraphics[width=10cm]{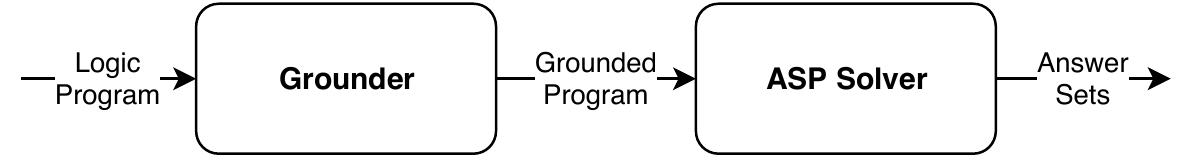}
	\caption{Typical ASP system architecture\label{fig:basic-arch}}
\end{figure}

\begin{figure}[t]
		\includegraphics[width=9cm]{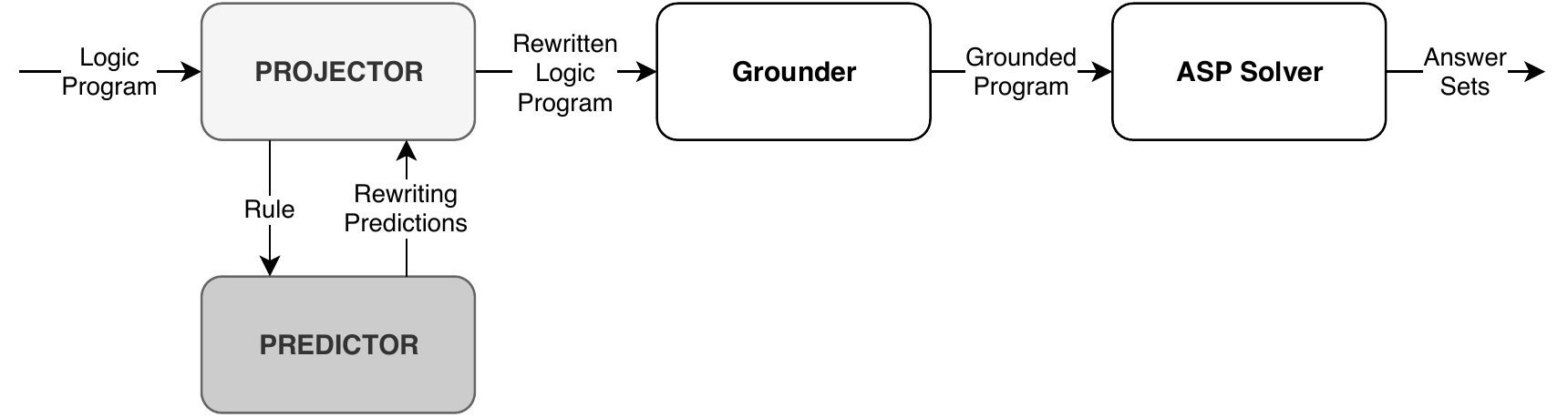}
	\caption{An ASP system  with \pr using \prd}
	\label{fig:pred-arch}
\end{figure}

Intuitive ASP encodings are not always the most optimal/performant, making this programming paradigm less attractive to novice users as their first attempts to problem solving may not scale.  ASP programs often require careful design and expert knowledge in order to achieve  performant results \citeb{gekakasc11a}.
Figure~\ref{fig:basic-arch} depicts a typical ASP system architecture.
The first step performed by systems   called grounders transforms a non-ground logic program (with variables) into a ground/propositional program (without variables). Expert ASP programmers often modify their ASP solution targeting the reduction of grounding size of a resulting program. Size of a ground program has been shown to be a predictive factor of a program's performance, enabling it to be used as an ``optimization metric'' \citeb{gekakasc11a}.
Intelligent grounding techniques~\citeb{fab12} utilized by grounders such as \gringo~\citeb{geb07b} or \idlv~\citeb{cal17}  also keep such a reduction in mind.
Intelligent grounding procedures analyze a given  (non-ground) program to produce a smaller propositional program without altering the solutions.
In addition, researchers looked into automatic program rewriting procedures.
Systems such as {\sc simplify}~(\cite{eit06}; \cite{eit06a}), \lpopt~(\cite{bic15}; \cite{bic16}), and \pr~\citeb{hippen2019automatic}  rewrite non-ground programs (preserving their semantics) targeting the reduction of the grounding size. 
These systems are meant to be prepossessing tools agnostic to the later choice of ASP solving technology.
Tools such as {\sc simplify}, \lpopt, and {\pr}, despite illustrating promising results, often hinder their objective. Sometimes, the original set of rules is better than the rewritten set, when their size of grounding and/or runtime is taken as a metric. Research has been performed to mitigate the negative impact of these rewritings. For example, \cite{mastria2020machine} demonstrated a novel approach to guide automatic rewriting techniques performed in \idlv using machine learning with a set of features built from structural properties of a considered program and domain information. Thus, a machine learning model guides \idlv on whether to perform built-in rewritings or not. 
Another example of incorporating automatic rewriting techniques with the use of information about specifics of a considered program and a considered grounder is work by \cite{calimeri2018optimizing}. In that work, the authors
incorporated program rewriting technique stemming from~\lpopt into the intelligent grounding algorithm of grounder~\idlv.
Such tight coupling of the rewriting and grounding procedures allows \idlv to make {a decision on whether to apply or not an \lpopt rewriting} based on the current state of grounding. Grounder \idlv accurately estimates the impact of rewriting on grounding and based on this information decides whether to perform a rewriting. This synergy of intelligent grounding and a rewriting technique demonstrates the best performant results. Yet, it makes the transfer of rewriting techniques laborious assuming the need of tight integration of any rewriting within a grounder of choice. {\em Here}, we propose an algorithm for estimating the size of grounding a program based on (i) mimicking an intelligent grounding procedure documented by~\cite{fab12}  and (ii) techniques used in query optimization in relational databases, see, for instance, Chapter~13 by \cite{silberschatz1997database}.
We then implement this algorithm in a system called \prd. This tool is meant to be used as a decision support mechanism for ASP program rewriting systems so that they perform a possible rewriting based on estimates produced by \prd.
This work culminates in the integration of  \prd within the rewriting tools \pr and \lpopt, which then are used prior to the invocation of a typical grounder-solver pair of ASP. For example, Figure~\ref{fig:pred-arch} depicts the use of \prd within the rewriting system \pr as a preprocessing step before the invocation of an ASP system.
To depict the use of \prd within the rewriting system \lpopt as a preprocessing step it is sufficient to replace the box named \pr by a box named \lpopt in Figure~\ref{fig:pred-arch}.
We illustrate the success of this synergy by an experimental analysis. It is due to note that \prd is a stand alone tool and can be used as part of any ASP inspired technology where its functionality is of interest.

We underline that the important contribution of this work is in the design of a building block -- in the shape of the system {\prd} -- towards making ASP a truly declarative framework.
Answer set programming is frequently portrayed as a powerful
declarative programming formalism. Yet, we can argue that such a claim is somewhat misleading.
At present, to achieve scalable ASP solutions to problems of interest, it is typical that {\em an expert ASP programmer} -- with strong insights into underlying grounding/solving technology --   constructs logic programs/encodings for problems that are efficient rather than intuitive. The ASP experts must rely on their extensive knowledge of the ASP technology to deliver efficient solutions. 
Yet, in truly declarative formalism we would expect the possibility of constructing {\em intuitive} encodings and rely on underlying systems to  process these efficiently.
This way programmers may  focus on coding specifications of problems at hand rather than the specifics of the shape of these specifications and the details of the underlying technology. This paper targets the development of infrastructure, which {\em one day} will allow us to achieve the ultimate goal of {\em truly declarative ASP}.
Ultimately, 
an expert ASP programmer capable of devising efficient encodings will be replaced by an ASP user capable of devising intuitive specifications that are then turned into effective specification by a portfolio of automatic tools such as, for example, \pr and \prd, or \lpopt and \prd pairs showcased and evaluated here in the final section of the paper. This work makes a step towards  achieving the described ultimate goal: it provides us with insights and possible directions for the developments on that pass.

\paragraph{Related work}
It is due to remark on another body of research that targets a similar goal namely portfolio-like approaches, where researchers use machine learning based methods in navigating the space of distinct ASP grounders and/or solvers -- {\sc claspfolio}~\citeb{hoo14}; {\sc me-asp}~\citeb{mar14};  or encodings -- {\sc esp}~\citeb{liu22}   to decide on the best  possibility in tackling considered problem by means of ASP technology.
All and all,  
to the best of our knowledge this work is \textit{one of the very few} approaches for the stated/similar purpose. Already mentioned work by \cite{mastria2020machine} presents an alternative machine learning based method for a similar purpose. In that work  properties of a program are considered to predict whether rewriting will help an ASP solver down the road or not.
Also, the work by \cite{calimeri2018optimizing} can be seen as the most related one to this paper. The greatest difference of the championed approach is its detachment from any specific grounding system. It produces its estimates looking at a program alone. \citeauthor{calimeri2018optimizing} incorporate computation of estimates within a grounder. The benefit of such approach that at any point in time their estimates are reflective of de facto grounding that happened so far.

\paragraph{Outline of the paper} We start by introducing the subject matter terminology. The key contribution of the work lies in the development of  formulas for estimating the grounding size of a logic program based on its structural analysis and insights on intelligent grounding procedures. First, we present the simplified version of these formulas for the case of tight programs. We trust that this helps the reader to build intuitions for the work. Second, the formulas for non-tight programs are given. We then describe the implementation details of system \prd. The main part of the presentation concerns most typical logic rules (stemming from Prolog). The section that follows the presentation of the key concepts discusses other kinds of rules and their treatment by the \prd system. We conclude by experimental evaluation that includes incorporation of \prd within rewriting systems \pr and \lpopt.

Parts of this paper appeared in the proceedings of the 17th Edition of the European Conference on Logics in Artificial Intelligence \citep{lie21a}.

\section{Preliminaries}\label{sec:prel}
An \emph{atom} is an expression $p(t_1,...,t_k)$, where $p$ is a predicate symbol of arity $k \geq 0$ and $t_1,...,t_k$ are {\em terms} -- either object constants or  variables. As customary in logic programming, variables are marked by an identifier starting with a capital letter. 
We assume object constants to be numbers. This is an inessential restriction as we can map strings to numbers using, for instance, the lexicographic order. 
For example, within our implementation described in this paper: we consider  all alphanumeric object constants occurring in a program; 
sort these object constants using the  lexicographic order; and map each string in this sorted list to a natural number that corresponds to its position in the list added to the greatest natural number occurring in the program.

For an atom $p(t_1,...,t_k)$ and position $i$ ($1\leq i\leq k$), we define an \emph{argument} denoted by $p[i]$. By  $p(t_1,...,t_k)^0$ and $p(t_1,...,t_k)^i$
we refer to predicate symbol $p$ and the term~$t_i$, respectively.
A \emph{rule} is an expression of the form
\beq
a_0 \ar a_1,...,a_m,not~a_{m+1},...,not~a_n.
\eeq{eq:rule}
where $n \geq m \geq 0$, $a_0$ is either an atom or symbol $\bot$, and $a_1,...,a_n$ are atoms. We refer to $a_0$ as the \emph{head} of the rule and an expression to the right hand side of an arrow symbol in~\eqref{eq:rule} 
as the \emph{body}. An atom $a$ and its negation $not~a$ is a {\em literal}. To literals $a_1,...,a_m$ in the body of rule~\eqref{eq:rule} we refer as \emph{positive}, whereas to literals $not~a_{m+1},...,not~a_n$ we refer as \emph{negative}. For a rule $r$, by $\mathbb{H}(r)$ we denote the head atom of~$r$. By $\mathbb{B}^+(r)$ we denote the set of positive literals in the body of $r$. We obtain the set of variables present in an atom~$a$ and a rule $r$ by $\vars(a)$ and  $\vars(r)$, respectively. 
For a variable~$X$ occurring in rule $r$, by $\args(r,X)$ we denote  the set 
$$
\{p[i]\mid a\in \mathbb{B}^+(r), a^0=p,
 \text{ and } a^i=X\}.
$$
In other words, $\args(r,X)$  denotes the set of arguments in the positive literals of rule $r$, where variable $X$ appears.  
A rule $r$ is \emph{safe} if each variable in $r$ appears in~$\mathbb{B}^+(r)$.
Let $r$ be a safe rule
\begin{align}
p(A) \ar q(A,B), r(1,A), not~s(B).\label{rule:safe}
\end{align}
Then $\vars(r)=\{A,B\}$,
$\args(r,A) = \{q[1], r[2]\}$, and
$\args(r,B) = \{q[2]\}$.
A \emph{(logic) program} is a finite set of safe rules. We call programs containing variables  \emph{non-ground}. 

For a program $\Pi$, $oc(p[i])$ denotes the set of all object constants occurring within $$
\{\mathbb{H}(r)^i \mid r\in \Pi \text{ and } \mathbb{H}(r)^0=p\},
$$ whereas $oc(\Pi)$ denotes the set of all object constants occurring in the head atoms of the rules in~$\Pi$. 

\begin{exmp}\label{examlpe1}
Let
 $\Pi_1$ denote a program
\begin{align}
&p(1).\ p(2).\ r(3). \label{rule:prog-1-gr-rules}\\
&q(X,1) \ar p(X). \label{rule:prog-1-rule}
\end{align}
Then,  $oc(p[1])=\{1,2\}$, 
$oc(q[1])=\emptyset$, $oc(q[2])=\{1\}$ and
$oc(\Pi_1)=\{1,2,3\}$.
The {\em grounding} of a program $\Pi$, denoted  $gr(\Pi)$, is a ground program obtained by instantiating variables in $\Pi$  with all object constants of the program. For example, $gr(\Pi_1)$ consists of rules in~\eqref{rule:prog-1-gr-rules} and rules
\begin{align}
&q(1,1) \ar p(1).~~~q(2,1) \ar p(2). \label{rule:gr-only1}\\
&q(3,1) \ar p(3). \label{rule:gr-only}
\end{align}
\end{exmp}

 Given a program $\Pi$, ASP grounders utilizing intelligent grounding are often able to produce a program smaller than its grounding $gr(\Pi)$, but that has the same answer sets as $gr(\Pi)$. 
 Recall program $\Pi_1$ introduces in Example~\ref{examlpe1}.
 For instance, the program obtained from $gr(\Pi_1)$ by dropping rule~\eqref{rule:gr-only} may be a result of intelligent grounding. 
The \emph{ground extensions} of a predicate within a  grounded program $\Pi$ are the set of terms associated with the predicate in the program. For instance,
in $gr(\Pi_1)$, the ground extensions of predicate  $q$ is the set of tuples
$
\{\langle 1,1 \rangle,\langle 2,1 \rangle, \langle 3,1 \rangle\}
$.
For an argument $p[i]$ and a  ground program $\Pi$, we 
call the number of distinct object constants occurring in the ground extensions of~$p$ in $\Pi$ at  position $i$ the \emph{argument size} of $p[i]$. For instance, for program $gr(\Pi_1)$ argument sizes 
of $p[1]$, $q[1]$, and $q[2]$ are $3$, $3$, and~$1$, respectively.

The \emph{dependency graph} of a program $\Pi$ is a directed graph $G_\Pi=\langle N,E \rangle$ such that~$N$ is the set of predicates appearing in $\Pi$ and $E$ contains the edge $(p,q)$ if there is a rule $r$ in $\Pi$ in which~$p$ occurs in~$\mathbb{B}^+(r)$ and $q$ occurs in the head of $r$.  A program $\Pi$ is \emph{tight} if $G_\Pi$ is acyclic, otherwise the program is \emph{non-tight}~\citeb{fag94}.

\begin{figure}[t]
\begin{center}
\begin{tabular}{lcr}
	\includegraphics[width=3cm]{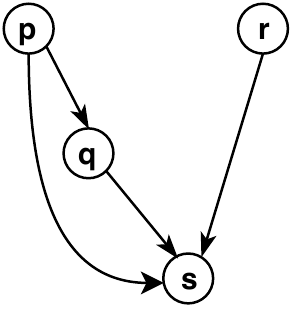}&
	\includegraphics[width=3cm]{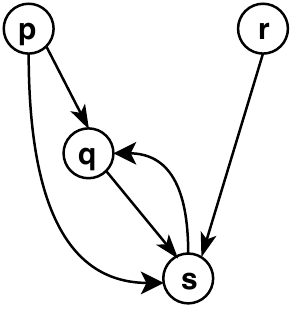}&
		\includegraphics[width=3cm]{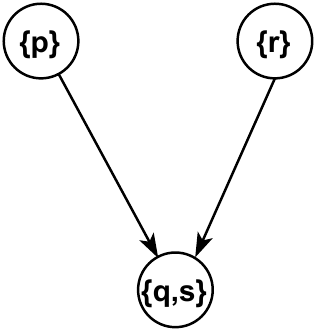}
	\end{tabular}
	\end{center}
	\caption{Left:  Graph $G_{\Pi_2}$; Center: Graph $G_{\Pi_3}$; Right: Graph $G^{sc}_{\Pi_3}$}

	\label{fig:dep-graph}
\end{figure}

\begin{exmp}\label{example2}
Let~$\Pi_2$ denote a program constructed from~$\Pi_1$ (introduced in Example~\ref{examlpe1}) by extending it with  rules: 
\begin{align}
&r(2).~ r(4).\label{rule:r24-def}\\
&s(X,Y,Z) \ar r(X), p(X), p(Y), q(Y,Z).\label{rule:s-def}
\end{align}
Program $\Pi_3$ is the program $\Pi_2$ extended with the rule:
\begin{align}
q(Y,X) \ar s(X,Y,Z).\label{rule:q-def-rec}
\end{align}
Figure~\ref{fig:dep-graph} shows the dependency graphs  $G_{\Pi_2}$ (left) and $G_{\Pi_3}$ (center). Program $\Pi_2$ is tight, while program~$\Pi_3$ is not.
\end{exmp}
\label{ex:pi2-pi3}

\section{System \prd }\label{sec:sol}
The key contribution of this work is the development of the system \prd (its algorithmic and software base), whose goal is to provide estimates for the size of an ``intelligently'' grounded program. In other words, its goal is to assess the impact of grounding without grounding itself.  \prd is based on the intelligent grounding procedures implemented by the grounder \dlv, described in~\cite{fab12}. The key difference is that, instead of building the ground instances of each rule in the program, \prd constructs statistics about the predicates, their arguments, and rules of the program. This section provides formulas we developed in order to produce the estimates backing up the computed statistics. We conclude with details on the implementation.

It is due to make couple remarks. First, in a way we parallel the work on query optimization techniques within relational databases, e.g., see Chapter 13 in \citeb{silberschatz1997database}. Indeed, when a particular query is considered within a relational database there are often numerous ways to its execution/implementation. Relational databases maintain statistics about its tables to produce estimates for intermediate results of various execution scenarios of potential queries. These estimates help database management systems decide which of the possible execution plans of the query at hand to select. In this work, we develop methods to collect and maintain  statistics/estimates about entities of  answer set programs. We then show how these estimates may help a rewriting (prepossessing) system for ASP  to decide whether to rewrite some rules of a program or not. 

Second,  the  intelligent grounding procedure implemented by grounder \dlv~\citeb{fab12}
is based on database evaluation techniques \citeb{ulm88,abi95}.
The same statement is the case for another modern grounder {\sc gringo}~\citeb{geb07b,kam22}. It also shares a lot in common with grounder {\sc dlv}. This fact makes the estimates of system \prd rooting in the algorithm of \dlv applicable also within the framework of {\sc gringo}. 
In a nutshell, both {\sc dlv} and {\sc gringo} 
instantiate a program via an iterative bottom-up process starting from the
program’s facts targeting the accumulation of ground atoms and ground rules 
derivable from the rules seen so far. As this process continues, a new ground rule is produced when its
positive body atoms belong to the already computed atoms. Then, the head atom of this rule is
added to the set of already accumulated ground atoms. This process continues until no new ground atoms/rules are produced by this process.


\medskip
\noindent
{\bf Argument size estimation}~
\underline{Tight program case:}~~
The estimation formulas are based on predicting argument sizes. To understand these it is essential to describe an order in which we produce estimates for predicate symbols/arguments. Given  a program~$\Pi$, we obtain such an ordering by performing a topological sorting on its dependency graph. We associate each node in this ordering with its position and call it a {\em strata rank} of a predicate. For example, $p,q,r,s$ is one possible ordering for program $\Pi_2$ (introduced in Example~\ref{example2}). This ordering  associates strata ranks $1,2,3,4$ with predicates $p,q,r,s$, respectively.


We now introduce some intermediate formulas for constraining our estimates. These intermediate formulas are inspired by query optimization techniques within relational databases, e.g., see Chapter 13 in \citeb{silberschatz1997database}.
These formulas keep track of information that helps us to estimate which actual values may occur in the grounded program without storing these values themselves. Let $p[i]$ be an argument. 
We track the range of values that may occur at this argument. To provide intuitions for an introduced process, consider an intelligent grounding of~$\Pi_2$ consisting of rules~(\ref{rule:prog-1-gr-rules}), (\ref{rule:gr-only1}), (\ref{rule:r24-def}), and rules
\begin{align}
&s(2,1,1) \leftarrow r(2), p(2), p(1), q(1,1).\label{rule:pi2-s1}\\
&s(2,1,1) \leftarrow r(2), p(2), p(2), q(2,1).\label{rule:pi2-s2}
\end{align}
This intelligent grounding
produces rules~\eqref{rule:pi2-s1},~\eqref{rule:pi2-s2} in place 
of rule~\eqref{rule:s-def}. Variable~$X$ from rule~\eqref{rule:s-def} is only ever replaced with  object constant~$2$. Intuitively, this is due to the intersection $oc(p[1]) \cap oc(r[1]) = \{2\}$. We model such a restriction by considering what minimum and maximum values are possible for each argument in an intelligently grounded program (compliant with described principle; all modern intelligent grounders respect such a restriction). We then use these values to define an ``upper restriction'' of the argument size for each argument.

\def\minte{\downarrow^{t\text{-}t}_{est}\!}
\def\maxte{\uparrow^{t\text{-}t}_{est}\!}
\def\rangete{range^{t\text{-}t}_{est}\!}
\def\ste{S^{t\text{-}t}_{est}}

For a tight program $\Pi$, let $p[i]$ be an argument in $\Pi$; $R$ be the following set of rules
\beq
\{r\mid r\in\Pi,~\text{$\mathbb{H}(r)^0=p$, and $\mathbb{H}(r)^i$ is a variable}\}.
\eeq{eq:setR}
By $\minte(p[i])$ we denote an estimate of a minimum value that may appear in argument $p[i]$ in  $\Pi$: 
	\begin{align*}
	&\minte(p[i])=min\big(oc(p[i])~ \cup\\
	& ~~~~\{max\Big( \{\minte(p'[i'])\mid p'[i']\in \args(r,\mathbb{H}(r)^i)\}\Big)\mid r \in R\}\big).
	\end{align*}
	The superscript \textit{t-t} stands for ``tight''. Note how $\mathbb{H}(r)^i$ in $\args(r,\mathbb{H}(r)^i)$ is conditioned to be a variable due to the choice of set $R$ of rules.
The function $\minte$ is total because the rank of the predicate occurring on the left hand side of the definition above is strictly greater than the ranks of all of the predicate symbols~$p'$ on the right hand side, where  rank is understood as a  strata rank defined before (multiple strata rankings are possible; any can be considered here).
By $\maxte(p[i])$ we denote 
an estimate of a maximum value that may appear in argument $p[i]$ in tight program $\Pi$. It is computed using formula for $\minte(p[i])$ with $min$, $max$, and $\minte$ replaced by $max$, $min$, and $\maxte$, respectively.

Now that we have estimates for minimum and maximum values, we estimate the size of the range of possible values.
We understand the \emph{range} of an argument to be the number of values we anticipate to see in the argument within an intelligently grounded program if the values were all integers between the minimum and maximum estimates.
It is possible that our minimum estimate for a given argument is greater than its maximum estimate. Intuitively, this indicates that no ground rule will contain this argument in its head. The number of values between the minimum and maximum estimates may also be greater than the number of object constants in a considered program. In this case, we restrict the range to the number of object constants occurring in the program. We compute the range, $\rangete(p[i])$, as follows:
\begin{align*}
&min\big(\{
max(\big\{0, \maxte(p[i]) - \minte(p[i]) + 1\big\}),|oc(\Pi)|
\}\big)
\end{align*}

\begin{exmp}
Recall program $\Pi_2$ introduced in Example~\ref{example2}. The operations required to compute the minimum estimate for argument~$s[1]$ in~$\Pi_2$ follow:
\begin{align*}
&\minte(r[1]) = min\big(oc(r[1])\big) = 2\\
&\minte(p[1]) = min\big(oc(p[1])\big) = 1\\
&\minte(s[1]) = min(oc(s[1]) \cup\\
&~~~\{max\big(\big\{\minte(r[1]), \minte(p[1])\big\}\big)\})
=min(\emptyset\cup\{2\})=2
\\
\end{align*}
We compute $\maxte(s[1])$ to be $2$. Then, $\rangete(s[1])$ is
\begin{align*}
&min(\{
max\big(\big\{0, \maxte(s[1]) - \minte(s[1]) + 1\big\}\big),|oc(\Pi_2)|
\})\\
&= min(\{
max\big(\big\{0, 2-2+1\big\}\big),4
\}) = 1
\end{align*}
\end{exmp}
We presented formulas for estimating the range of values in program's arguments. We now show how these estimates are used to assess the \emph{size} of an argument understood as the number of distinct values occurring in this argument upon an intelligent grounding.  
We now outline intuitions behind a recursive process that we capture in formulas.
Let $p[i]$ be an argument.  If $p[i]$ is such that predicate $p$ has no incoming edges in the program's dependency graph, then we  estimate  the size of $p[i]$ as $|oc(p[i])|$. Otherwise, consider  rule~$r$ such that $\mathbb{H}(r)^0=p$ and $\mathbb{H}(r)^i$ is a variable. Our goal is to estimate the \emph{number of values} variable $\mathbb{H}(r)^i$ may be replaced with during intelligent grounding. To do so, we consider the argument size estimates for arguments in the positive body of the rule that contain variable $\mathbb{H}(r)^i$. Based on  typical intelligent grounding procedures,  variable $\mathbb{H}(r)^i$ may not take  more values than the minimum of those argument size estimations. This gives us an estimate of the argument size relative to a single rule $r$. The argument size estimate of $p[i]$ with respect to the entire program may be then  computed as the sum of such estimates for all rules such as $r$ (recall that rule $r$  satisfies the requirements $\mathbb{H}(r)^0=p$ and $\mathbb{H}(r)^i$ is a variable). 
Yet, the sum over all rules may heavily overestimate the argument size. To lessen the effect of overestimation we incorporate range estimates discussed before into the described computations.

For a tight program $\Pi$, let $p[i]$ be an argument in $\Pi$; 
$R$ be the set~\eqref{eq:setR} of rules.
By $\ste(p[i])$ we denote an estimate of the argument size $p[i]$ in $\Pi$. This estimate is computed as follows:
	\begin{align*}
	&\ste(p[i])=min\Big(\Big\{\rangete(p[i]),~|oc(p[i])| + \\
	&~~~~\sum_{r\in R}min\big(\{\ste(p'[i'])\mid p'[i']\in \args(r,\mathbb{H}(r)^i)\}\big)\Big\}\Big)
	\end{align*}
We can argue that the function $\ste$ is total in the same way as we argued that the function 
$\minte$ is total.

\begin{exmp}
Let us illustrate the computation of the argument size estimates for argument $s[2]$ in program~$\Pi_2$ (introduced in Example~\ref{example2}). Given
that~${\rangete(s[2]) = 2}$ and $oc(s[2])=\emptyset$:
\begin{align*}
&\ste(p[1]) = |oc(p[1])| = 2\\
&\ste(q[1]) = min(\rangete(q[1]),\{|oc(q[1])| +\\
&~~~min\big(\{\ste(p[1])\}\big)\}) = min(\{2,0 + min(\{2\})\}) = 2\\
&\ste(s[2]) = min\big(\rangete(s[2]), \\
&~~~ 
\big\{|oc(s[2])| + min\big(\{\ste(p[1]), \ste(q[1])\}\big) 
\big\}\big)  = 2
\end{align*}
\end{exmp}

\medskip
\noindent
\underline{Arbitrary (nontight) program case:}~~
To process arbitrary programs (tight and non-tight), we must manage  the circular dependencies such as present in sample program~$\Pi_3$ defined in Example~\ref{example2} in the section on preliminaries. We borrow and simplify a concept of the component graph by \cite{fab12}.
The \emph{component graph} of a program $\Pi$ is an acyclic directed graph $G^{sc}_\Pi = \langle N,E \rangle$ such that~$N$ is the set of strongly connected components in the dependency graph $G_\Pi$ of $\Pi$ and $E$ contains the arc $(P,Q)$ if there is an arc $(p,q)$ in $G_\Pi$ where $p \in P$ and $q \in Q$. For tight programs, we identify its  component graph with the dependency graph itself by associating a singleton set annotating a node with its member. Figure~\ref{fig:dep-graph} (right) shows the component graph for program $\Pi_3$.
For  a program~$\Pi$, we obtain an ordering on its predicates by performing a topological sorting on its component graph. We associate each node in this ordering with its position and call it a {\em strong strata rank} of each predicate that belongs to a node. For example, $\{p\},\{r\},\{q,s\}$ is one possible topological sorting of $G^{sc}_{\Pi_3}$. This ordering  associates the following strong strata ranks $1,2,3,3$ with predicates $p,r,q,s$, respectively.

Let $C$ be a node/component in graph $G^{sc}_\Pi$. By $\mathcal{P}_C$ we denote~the set
$$\{r\mid p\in C, r\in\Pi,\text{~and~} \mathbb{H}(r)^0=p\}.$$ 
We call this set  a \emph{module}. A rule $r$ in module $\mathcal{P}_C$ is a \emph{recursive rule} if there exists an atom $a$ in the positive body of $r$ so that $a^0=p$ and predicate $p$ occurs  in $C$. Otherwise, rule $r$ is an \emph{exit rule}.
For tight programs, all rules are exit rules. 
It is also possible to have modules with only recursive rules.

\begin{exmp}
The modules in program $\Pi_3$ introduced in Example~\ref{example2} contain
\begin{align*}
&\mathcal{P}_{\{p\}} = \{p(1).~~~p(2).\};~~\mathcal{P}_{\{r\}} = \{r(2).~~~r(3).~~~r(4).\};
\end{align*}
and $\mathcal{P}_{\{q,s\}}$ composed of rules~(\ref{rule:prog-1-rule}),~(\ref{rule:s-def}), and~(\ref{rule:q-def-rec}).
The  rules rules~(\ref{rule:s-def}) and (\ref{rule:q-def-rec}) are recursive.
\end{exmp}


In the sequel we consider components whose module contains an exit rule. For a component~$C$ and its module~$\mathcal{P}_C$, we construct a partition $M_1,...,M_n$ ($n \geq 1$) in the following way:
Every exit rule of $\mathcal{P}_C$ is a member of $M_1$.
A recursive rule $r$ in $\mathcal{P}_C$ is a member of~$M_{k}$ ($k>1$) if 
\begin{itemize}
\item for every predicate $p\in C$ occurring in $\mathbb{B}^+(r)$, there is a rule $r'$ in $M_1 \cup...\cup M_{k-1}$, where $\mathbb{H}(r')^0=p$ 
and 
\item there is a predicate $q$ occurring in $\mathbb{B}^+(r)$ such that there is a rule $r''$ in $M_{k-1}$, where  $\mathbb{H}(r'')^0=q$. 
\end{itemize}
We refer to the unique partition created in this manner as the \emph{component partition} of $C$; integer~$n$ is called its {\em cardinality}. We call elements of a component partition \emph{groups} (the component partition is undefined for components whose module does not contain an exit rule).
Prior to illustrating these concepts by an example we introduce one more notation.
For a component partition $M_1,\dots,M_k,\dots,M_n$, by 
 $M^{p[i]}_k$ 
we denote  the set 
$$\{r\mid r\in M_k,~\text{$\mathbb{H}(r)^0=p$, and $\mathbb{H}(r)^i$ is a variable}\};$$
 and by $M^{p[i]}_{1...k}$ we denote  the~union~$\bigcup_{j=1}^k M^{p[i]}_{j}$.

\begin{exmp}
Recall program~$\Pi_3$ from Example~\ref{example2}.
The component partition of node $\{q,s\}$ in $G^{sc}_{\Pi_3}$  follows:
$$
\ba{l}
M_1=\{q(X,1) \leftarrow p(X).\}\\
M_2=\{s(X,Y,Z) \leftarrow r(X), p(X), p(Y), q(Y,Z).\}\\
M_3=\{q(Y,X) \leftarrow s(X,Y,Z).\}.
\ea
$$
For program~$\Pi_3$ and its argument~$q[1]$:
\begin{align*}
M^{q[1]}_{1...3} = \{q(X,1) \leftarrow p(X).~~~q(Y,X) \leftarrow s(X,Y,Z).\}
\end{align*}
\end{exmp}

We now generalize  range and argument size estimation formulas for tight programs to the case of arbitrary programs. These formulas are more complex than their ``tight versions'', yet they perform similar operations at their core. Intuitively, formulas for tight programs rely on argument ordering provided by the program's dependency graph. Now, in addition to an order provided by the component dependency graph, we rely on the orders given to us by the component partitions of the program.

 \def\mine{\downarrow_{est}\!\!}
\def\maxe{\uparrow_{est}\!\!}

 \def\minge{\downarrow^{gr}_{est}\!\!}
\def\maxge{\uparrow^{gr}_{est}\!\!}

 \def\minre{\downarrow^{rule}_{est}\!\!}
\def\maxre{\uparrow^{rule}_{est}\!\!}
 \def\minse{\downarrow^{split}_{est}\!\!}
\def\maxse{\uparrow^{split}_{est}\!\!}

\def\rangee{range_{est}\!}
\def\se{S_{est}}
\def\sge{S^{gr}_{est}\!}
\def\sre{S^{rule}_{est}\!}
\def\sse{S^{split}_{est}\!}

In the remainder of this section, let
$\Pi$ be a program;
 $p[i]$ be an argument in $\Pi$; $C$ be the node in the component graph of $\Pi$ so that $p\in C$; 
 $n$ be the cardinality of the component partition of~$C$; and $j$ be an integer such that $1 \leq j \leq n$.
 
 If the module of $C$ does not contain an exit rule, then
 the estimate of the range of an argument~$p[i]$, denoted $\rangee(p[i])$, is assumed $0$ and  the estimate of the size of an argument $p[i]$, denoted $\se(p[i])$,  is assumed $0$.

We now consider the case when the module of $C$ contains an exit rule.
By~$\mine(p[i])$ we denote an estimate of a minimum value that may appear in argument $p[i]$ in  program~$\Pi$:
\begin{align*}
&\mine(p[i]) = \minge(p[i], n)\\
&\minge(p[i], j) =min (oc(p[i])\cup\{\minre(p[i], j, r)\mid r \in M^{p[i]}_{1...j}\}
)\\
&\minre(p[i], j, r) =max\big(\big\{\minse(p[i], p'[i'], j)\mid p'[i'] \in \args(r,\mathbb{H}(r)^i)\big\}\big)\\
&\minse(p[i], p'[i'], j) =\begin{cases}
\minge(p'[i'], j-1), & \text{if}\ p'\ \text{in the same component as}\ p\\
\mine(p'[i']), & \text{otherwise}
\end{cases}
\end{align*}
\\
We note the strong similarity between the combined definitions of $\minge(p[i],j)$ and $\minre(p[i],j,r)$ compared to the corresponding ``tight'' formula~$\minte(p[i])$. Formula for \hbox{$\minse(p[i],p'[i'],j)$} serves two purposes. 
If the predicate $p'$ is in the same component as predicate~$p$, we decrement the counter $j$ (intuitively bringing us to preceding groups in component partition).  Otherwise, we simply use the minimum estimate for $p'[i']$ that is due to the computation relevant to another component.

We now show that defined functions $\mine$\;, $\minge$\;, $\minre$\; and $\minse$ are total. Consider any strong strata ranking of program's predicates. Then,  by $rank(p)$ we refer to the corresponding strong strata rank of a predicate $p$. The following table provides ranks associated with expressions used to define functions in question:

\vspace{2mm}
 \begin{tabular}{l|l}
Expression&~Rank\\
 $\mine(p[i])$& ~$\omega\cdot (rank(p)+1)$\\
  $\minge(p[i], j)$ &~  $\omega\cdot rank(p)+j$\\
$\minre(p[i], j, r)$&~ $\omega\cdot rank(p)+j$\\
$\minse(p[i], p'[i'], j)$&~  $\omega\cdot rank(p)+j$\\
\end{tabular}\\
\vspace{2mm}

\noindent
where $\omega$ is the smallest infinite ordinal number. It is easy to see that in definitions of  functions $\mine$\;, $\minge$\;, and $\minre$\; the ranks associated with their expressions do not increase. In definition of
$\minse$ in terms of $\mine$, the rank decreases. Thus, the defined functions are total.

By $\maxe(p[i])$ we denote 
an estimate of a maximum value that may appear in argument $p[i]$ in  program $\Pi$. It is computed using formula for $\mine(p[i])$ with $min$, $max$, 
$\mine$\;, $\minge$\;, $\minre$\;, and $\minse$ 
 replaced with $max$, $min$,
 $\maxe$\;, $\maxge$\;, $\maxre$\;, and $\maxse$\;, respectively. 
The range of an argument~$p[i]$, denoted $\rangee(p[i])$, is computed by the formula of $\rangete(p[i])$, where we replace $\minte$\; and $\maxte$\; with $\mine$\; and $\maxe$\;, respectively.  

We define the formula for finding the argument size estimates, $\se(p[i])$, as follows:
\begin{align*}
&\se(p[i]) = \sge(p[i], n)\\
&\sge(p[i], j) =min\big(\big\{\rangee(p[i]),|oc(p[i])| + \sum_{r \in M^{p[i]}_{1...j}} \sre(p[i], j, r)\big\}\big)\\
&\sre(p[i], j, r) =~min\big(\big\{\sse(p[i], p'[i'], j)~|~p'[i'] \in \args(r,\mathbb{H}(r)^i)\big\}\big)\\
&\sse(p[i], p'[i'], j) =\begin{cases}
\sge(p'[i'], j-1), & \text{if}\ p'\ \text{is in the same component as}\ p\\
\se(p'[i']), & \text{otherwise}
\end{cases}
\end{align*}
We can argue that the function $\se$ is total in the same way as we argued that the function 
$\mine$\; is total.

\smallskip
\noindent
{\bf Program size estimation}
~\underline{Keys}~~~ We borrow the concept of a key from relational databases. 
This concept allows us to produce more accurate final estimates as it carries important structural information about predicates and the kinds of instantiations possible for them. (Table~\ref{tab:keys} presented in the section on experimental analysis illustrates the impact of information on the keys within the implemented system.)
For some predicate~$p$, we refer to any set of arguments of~$p$ that can uniquely identify all ground extensions of~$p$ as a \emph{superkey} of~$p$. 
We call  a minimal superkey 
a {\em candidate key}. For instance, let 
 the following be the ground extensions of some predicate~$q$:
\begin{align*}
\{&\langle 1,1,a \rangle, \langle 1,2,b \rangle, \langle 1,3,b \rangle,
\langle 2,1,c\rangle, \langle 2,2,c\rangle, \langle 2,3,a\rangle\}
\end{align*}
It is easy to see that both $\{q[1],q[2]\}$ and $\{q[1],q[2],q[3]\}$ are superkeys of $q$, while $\{q[1]\}$ is not a superkey.
 Only superkey $\{q[1], q[2]\}$ is a candidate key. A \emph{primary key} of a predicate~$p$ is a single chosen candidate key. A predicate may have at most one primary key. For the purposes of this work, we allow the users of \prd to manually specify the primary key.  It is possible that some predicates do not have primary keys specified. To handle such predicates, we define $key(p)$ to mean the following:
\begin{align*}
key(p) =
\begin{cases}
\text{the primary key of}\ p, &  \text{if}\ p\ \text{has a primary key specified}\\
\{p[1],...,p[n]\}, & \text{otherwise}
\end{cases}
\end{align*}
where $n$ is the arity of $p$. We call an argument $p[i]$ a \emph{key argument} if it is in $key(p)$.
For a rule $r$, by $\kvars(r)$ we denote the set of its variables that occur  in its key arguments. 

\medskip

\noindent
\underline{Rule size estimation}~~~
We now have all the ingredients to provide an estimate for grounding size of each rule in a program. We understand a \emph{grounding size} of a rule as the number of rules produced as a result of intelligently grounding this rule.
For a rule $r$ in a program $\Pi$, the estimated grounding size, denoted
 $\se(r)$, is computed as follows:
\begin{align*}
\se(r) =
\prod_{X\in \kvars(r)}min\big(\{ \se(p[i])\mid p[i]\in \args(r,X)\}\big)
\end{align*}

\noindent
{\bf Implementation Details}
System \prd\footnote{\url{https://www.unomaha.edu/college-of-information-science-and-technology/natural-language-processing-and-knowledge-representation-lab/software/predictor.php}} is developed using the Python 3 programming language.
\prd utilizes \pyclingo version~5, a Python API sub-system of answer set solving toolkit \clingo~\citeb{gebser2015potassco}. The \pyclingo API enables users to easily access and enhance  ASP processing steps within Python code, including access to some data in the processing chain. In particular, \prd uses \pyclingo to parse a logic program into an abstract syntax tree (AST) representation. After obtaining the AST, \prd has an immediate access to internal rule structure of the program and computes estimates for the program using the presented formulas.
System \prd is designed for integration with other systems processing ASP programs. It is distributed as a package that can be imported into other systems developed in Python 3, or it can be accessed through a command line interface. 
In order to ensure that system \prd is applicable to real world problems, it supports  ASP-Core-2 logic programs. For instance, the estimation formulas presented here generalize well to programs with choice rules and disjunction. Rules with aggregates are also supported.  Yet, for such rules more sophisticated approaches are required to be more precise at estimations. Next section covers key details on the ASP-Core-2 support by the \prd system. We then conclude by integrating the \prd system into two rewriting tools, namely, \pr and \lpopt. We present a thorough experimental analysis for these systems and the enhancement that \prd offers to them.

\section{Language Extensions: ASP-Core-2 Support}\label{sec:ext-impl}
In order to ensure that system \prd is applicable to real world problems, it has been designed to operate on many common features of ASP-Core-2 logic programs. In the following we extend the definition of logic rules to include these features and discuss how these features are handled by \prd.

\paragraph{Pools and Intervals}
In ASP-Core-2 logic programs, an atom may have the form $p(t_1;...;t_n)$, where $p$ is a predicate of arity $1$, and $t_1;...;t_n$ is a semicolon separated list of terms. Here, $t_1;...;t_n$ is a \emph{pool} term. A predicate with a pool term is ``syntactic sugar'' that indicates there is a copy of that rule for every object constant in the pool.
\begin{exmp}
The following rule containing pool terms:
\begin{align*}
p(a;b) \leftarrow q(c;d).
\end{align*}
can be expanded to the following rules:
\begin{align*}
&p(a) \leftarrow q(c).\\
&p(a) \leftarrow q(d).\\
&p(b) \leftarrow q(c).\\
&p(b) \leftarrow q(d).
\end{align*}
\end{exmp}

Similarly, ASP-Core-2 programs may contain atoms of the form $p(l..r)$, where $p$ is a predicate of arity $1$, and $l$, $r$ are terms. Here, $l..r$ is an \emph{interval} term. A predicate with an interval term is ``syntactic sugar'' indicating that there is a copy of this rule for every integer between the range of $l$ to $r$, inclusive.
\begin{exmp}
The following rule containing interval terms:
\begin{align*}
p(1..3, a) \leftarrow q(1..2).
\end{align*}
can be expanded to the following rules:
\begin{align*}
&p(1,a) \leftarrow q(1).\\
&p(1,a) \leftarrow q(2).\\
&p(2,a) \leftarrow q(1).\\
&p(2,a) \leftarrow q(2).\\
&p(3,a) \leftarrow q(1).\\
&p(3,a) \leftarrow q(2).
\end{align*}
\end{exmp}

For both pool and interval terms, system \prd handles the program as though it were in its expanded form.

\paragraph{Aggregates}
An \emph{aggregate element} has the form
\begin{align*}
t_0,...,t_k : a_0,...,a_m, not~a_{m+1},..., not~a_n.
\end{align*}
where $k \geq 0$, $n \geq m \geq 0$, $t_0,...,t_k$ are terms and $a_0,...,a_n$ are atoms. An \emph{aggregate atom} has the form
\begin{align*}
\#aggr\{e_0,...,e_n\} \prec t
\end{align*}
where $n \geq 0$ and $e_0,...,e_n$ are aggregate elements. Symbol $\#aggr$ is either $\#count$, $\#sum$, $\#max$, or $\#min$. Symbol $\prec$ is either $<$, $\leq$, $=$, $\neq$, $>$, or $\geq$. Symbol $t$ is a term.

System \prd supports rules containing aggregates to a limited extent. In particular, \prd will simplify such a rule as if it had no  aggregate atoms.
\begin{exmp}
The rule containing an aggregate atom:
\begin{align*}
p(X) \leftarrow q(X), \#count\{Y : r(X,Y)\} < 3.
\end{align*}
is seen by \prd as the following rule:
\begin{align*}
p(X) \leftarrow q(X).
\end{align*}
while the only variable seen in this rule will be $X$.
\end{exmp}

It is important to note that if an aggregate contains variables, it is possible that the \emph{length of a rule} expands during grounding processes, where it is understood that the length of a rule is the number of atoms in a rule. We do not consider this length expansion when computing the grounding size of a rule.

\paragraph{Disjunctive and Choice Rules}
A \emph{disjunctive rule} is an extended form of ASP logic rule that allows disjunctions in its head. They are of the form
\begin{align*}
a_0 \lor ... \lor a_k \leftarrow a_{k+1}, ..., a_m, not~a_{m+1}, ..., not~a_n.
\end{align*}
where $n \geq m \geq k \geq 0$, and $a_0,...,a_n$ are atoms.

System \prd handles a disjunctive rule by replacing it with the set of rules created in the following way. For each atom $a$ in the head of a disjunctive rule $r$, \prd creates a new rule of the form $a \leftarrow \mathbb{B}(r)$. For computing range and argument size estimates, all of these newly created rules are used. However, when estimating the grounding size of the original rule, only one of the rules is used.
\begin{exmp}
The disjunctive rule $r$:
\begin{align*}
p(1) \lor p(2) \leftarrow q(1).
\end{align*}
is replaced by the following two rules:
\begin{align*}
p(1) \leftarrow q(1).\\
p(2) \leftarrow q(1).
\end{align*}
Yet, only one of those rules is used for estimating the grounding size of the original rule. Using these rules is sufficient for estimating grounding information, even though they are not semantically equivalent to the original disjunctive rule.
\end{exmp}

A \emph{condition} is of the form
$$
a_0 : a_1,...,a_m, not~a_m+1,...,not~a_n
$$
where $n \geq m \geq 0$, and $a_0,...,a_n$ are atoms. We refer to $a_0$ as the head of the condition.
A \emph{choice atom} is of the form $l \{c_1 ; ... ; c_n\} r$, where $l$ is an integer, $r$ is an integer such that $r \geq l$, and $c_1;...;c_n$ is a semi-colon separated list of conditions. We now extend the definition of a rule given by~\eqref{eq:rule} to allow the head to be a choice atom. We refer to rules whose head contains a choice atom as \emph{choice rules}.

System \prd handles a choice rule similarly to the case of a disjunctive rule, replacing it with the set of rules created in the following way. For each atom $a$ in the head of a condition in the choice atom in rule~$r$, create a new rule of the form $a \leftarrow \mathbb{B}(r)$. For computing range and argument size estimates, all of these newly created rules are used. However, when estimating the grounding size of the original rule, only one of the rules will be used. Note that, as with aggregates, choice rules can increase the length of a rule.
\begin{exmp}
	The choice rule:
	\begin{align*}
	1\{p(X) : q(1); p(Y) \}1 \leftarrow r(X,Y), s(Y).
	\end{align*}
	is replaced by the following two rules:
	\begin{align*}
	p(X) \leftarrow r(X,Y), s(Y).\\
	p(Y) \leftarrow r(X,Y), s(Y).
	\end{align*}
	Yet, only one of those rules is used for estimating the grounding size of the original rule.
\end{exmp}

\paragraph{Functions}
In ASP-Core-2, a term may also be of the form $f(t_1,...,t_n)$, where $f$ is a function symbol and $t_1,...,t_n$ ($n>0$) are term. We call terms of this form \emph{function} terms. In order to be more compliant with ASP-Core-2 features, \prd is capable of running on programs containing function terms, however when a function term is encountered by \prd, it simply sees the function term as an object constant.

\paragraph{Binary Operations}
The ASP-Core-2 standard also allows \emph{binary operation} terms. A binary operation term is of the form $t_1~op~t_2$, where $t_1$ and $t_2$ are either an integer object constant, a variable, or a binary operation and $op$ is a valid binary operator\footnote{\url{http://potassco.sourceforge.net/doc/pyclingo/clingo.ast.html\#BinaryOperator}}. If an atom contains a binary operation term, system \prd handles it in one of three ways. If the binary operation has no variables, it treats the term as an object constant. If the binary operation contains exactly one variable, it treats the term as that variable. Otherwise, the atom is treated as if it were part of the negative body (and therefore not used in estimations).
\begin{exmp}
In the following rule containing binary operation terms:
\begin{align*}
\leftarrow p(1+1), q(2*X+1), r(2*X+Y), s(Y).
\end{align*}
the atoms are viewed as follows. Atom $p(1+1)$ is seen as containing an object constant term. Atom $q(2*X+1)$ is seen as the atom $q(X)$. Atom $r(2*X+Y)$ is seen as being part of the negative body.
\end{exmp}

\section{Experimental Analysis}\label{sec:proj-int}

We investigated the utility of system \prd by integrating it as a decision support mechanism into the ASP rewriting tool \pr to create tool \prdpr, as well as the ASP rewriting tool \lpopt, to create tool \prdlpopt. These tools are discussed in following subsections. 

\subsection{System \prdpr}
Figure~\ref{fig:pred-arch} (presented in the Introduction section) demonstrates how \prd is integrated with system \pr resulting in what we call \prdpr. Note how \prd runs entirely independent of and prior to the grounding step of a considered ASP grounder-solver pair. 

The rewriting tool~\pr is documented by~\cite{hippen2019automatic}. This tool focuses on so called projection technique. In its default settings, it studies each rule of a given program and when a projection rewriting is applicable to considered rules \pr rewrites these accordingly. Thus, whenever the rewriting is established to be possible it is also performed. The  \prdpr tool extends the \pr system by the decision making mechanism supported by \prd on whether to perform rewriting or not. When \pr establishes that a rewriting is possible the system \prd evaluates an original rule against its rewritten counterpart as far as their predicted grounding sizes. The projection rewriting will only be applied if the rewritten rule is predicted to produce smaller grounding footprint.
In particular,
for each rule $r$ in program $\Pi$, \pr will create a set $R$ of rules, which represents one of the possible ``projected''-versions of $r$. This set~$R$ of rules is then substituted into $\Pi$ to create program $\Pi'$. If the predicted grounding size for this new program is smaller than, or equal to the original, the set~$R$ of rules  is kept and $\Pi'$ becomes a   considered program in the future evaluations. However, if the new predicted grounding size is larger than the original, set $R$ is discarded, and \prdpr will move on to the next rule in $\Pi$. 
To summarize, tool \prd is used by \pr in two ways:
\vspace{1mm} 
\begin{enumerate}
    \item When \prdpr encounters a tie through its default heuristics of \pr for selecting  variables to project, \prdpr generates the resulting projections for each of the variables and use the projection that is predicted to have the smallest grounding size. 
    
    \item \prdpr only performs a projection if the prediction for the projection is smaller than the predicted grounding size for the original rule. 
\end{enumerate}
\vspace{1mm} 

We note that it is possible for projections to occur inside of aggregate expressions. System \prd is not used to decide if these projections should be performed, so that these projections always occur in \prdpr.


\subsection{System \prdlpopt}

Figure~\ref{fig:pred-arch} with the box representing \pr replaced by the box representing \lpopt demonstrates how \prd is integrated with system \lpopt. We refer to the version of \lpopt integrated with \prd as \prdlpopt. Once again, \prd runs entirely independent of and prior to the grounding step. 

The rewriting tool~\lpopt is documented by~\cite{bic15,bic16}. This tool focuses on so called rule decomposition technique. This technique is strongly related to a rewriting championed by  system \pr. In fact, \pr and \lpopt can be characterized as the tools performing the same kind of rewriting, while using different heuristics on how and when to apply this rewriting. Both systems attempt reducing the number of variables occurring in a rule by (a) introducing an auxiliary predicate and (b) replacing an original rule by new rules. In other words, there are often multiple ways available for rewriting the same rule and these systems may champion different ways. In its default settings, \lpopt studies each rule of a given program and when a rule decomposition rewriting is applicable to considered rules \lpopt rewrites these accordingly. Thus, it behaves just as the \pr system when used with its default settings: whenever the rewriting at hand is established to be possible it is also performed. The  \prdlpopt tool extends the \lpopt system by the decision making mechanism of \prd on whether to perform rewriting or not in the same manner as \prdpr tool extends the \pr system by the decision making mechanism  of \prd. We refer the reader to the previous subsection for the details.


\subsection{Evaluation}
To evaluate the usefulness of \prd, two sets of experiments are performed.
First, an ``intrinsic''  evaluation over accuracy of the predicted grounding size compared to
the actual grounding size is examined. 
Second, an  ``extrinsic'' evaluation of systems \prdpr and \prdlpopt is conducted to examine whether the system \prd is indeed of use as a decision support mechanism on whether to perform or not the rewritings of \pr and \lpopt, respectively. We note that the later evaluation is of a special value illustrating the value and the potential of system \prd and technology of the kind. It assesses \prd's impact when it is used in practice for its intended purpose as a decision making assistant. 
The intrinsic evaluation has its value in identifying potential future work directions and pitfalls in estimations. Overall, we will observe intrinsically that our estimates differ frequently in order of magnitude from the reality. Yet, extrinsic evaluation clearly states that  \prd performs as a solid decision making assistant for the purpose of improving rewriting tools when their performance depends on a decision when rewriting should take place versus not.


Benchmarks were gathered from two different sources. First, programs from the Fifth Answer Set Programming Competition \citeb{aspcomp5} were used. Of the 26 programs in the competition, 13 were selected (these that system \pr, in its default settings, has preformed rewritings on).
For each program, the {\bf 20} instances 
(originally selected for the competition) 
were used.
One interesting thing to note about these encodings is that they are generally already well optimized. As such, performing projections often leads to an increase in grounding size.
Second, benchmarks were gathered from an application called \aspccg implementing a natural language parser \citeb{aspccg11}. This domain has been extensively studied  in \cite{bud15} and was used to evaluate system \pr by \cite{hippen2019automatic}. In that evaluation, the authors considered 3  encodings from \aspccg: \enco, \encm, \encl.
We introduced changes to the encodings \enco, \encm, and \encl to make these in ASP-Core-2 standard~\cite{cal20} compatible with the \lpopt system.
We utilize the same  {\bf 60} instances as in the mentioned evaluation of  \pr.
In our experiments, system \pr was provided with the key information for some
root predicate arguments within several of the benchmarks.  Non-default keys used for all benchmarks can be found in Table~\ref{tab:keys}. The sign ``-'' within the table denotes benchmarks where no key information was provided by the user.

\begin{table}[h]
	\centering
	\caption{Key information for benchmark programs}
	\begin{tabular}{ll}
		\hline
		\textbf{Program}             & \textbf{Keys} \\ \hline
		Bottle Filling               & -               \\
		Hanoi Tower                  & -                \\
		Incremental Scheduling       & $\tt precedes/2[1]$, $\tt importance/2[1]$, $\tt job\_device/2[1]$,\\&$\tt job\_len/2[1]$, $\tt deadline/2[1]$, $\tt curr\_job\_start/2[1]$,\\&$\tt curr\_on\_instance/2[1]$, $\tt instances/2[1]$              \\
		Knight Tour with Holes       & -             \\
		Labyrinth                    & -             \\
		Minimal Diagnosis            & $\tt obs\_elabel/3[1,2]$ \\
		Nomystery                    & $\tt at/2[1]$, $fuel/2[1]$, $\tt goal/2[1]$ \\
		Permutation Pattern Matching & $\tt t/2[1]$, $\tt p/2[1]$\\
		Ricochet Robots              & $\tt amo/2[1]$, $\tt d1/2[1]$, $\tt dir/2[1]$ \\
		Solitaire                    & - \\
		Stable Marriage              & $\tt manAssignsScore/3[1,2]$, $\tt womanAssignsScore/3[1,2]$              \\
		Valves Location              & $\tt dem/3[1,2]$        \\
		Weighted-Sequence            & $\tt leafWeightCardinality/3[1]$ \\
		\aspccg~\enco;~\encm;~\encl                 &  $\tt word\_at/2[2]$, $\tt category\_tag\_nofeatures/3[1]$,\\&$\tt category\_tag/3[1]$, $\tt adjacent/2[1]$     \\
		\hline
	\end{tabular}
	\label{tab:keys}
\end{table}

Table~\ref{tab:prog-details} details interesting features in the programs from both domains. The second column provides information about some features present in the programs. These features are abbreviated with the meanings as follows (abbreviation letters bolded): \textbf{n}on-tight program, \textbf{a}ggregates, \textbf{b}inary operation terms, \textbf{c}hoice rules, and \textbf{f}unction terms. The competition benchmarks also consisted of two encodings: a newer 2014 encoding and a 2013 encoding from the previous year. The third column specifies which encoding was used (in case the newer encoding consisted of no projections).

\begin{table}[h]
	\centering
	\caption{Feature and version details for benchmark programs}
	\begin{tabular}{lll}
		\hline
		\textbf{Program}             & \textbf{Features} & \textbf{2013} \\ \hline
		Bottle Filling               & a,b                & \textbf{Yes}   \\
		Hanoi Tower                  & b                 & No             \\
		Incremental Scheduling       & a,b,c              & No             \\
		Knight Tour with Holes       & n,b              & No             \\
		Labyrinth                    & n              & No             \\
		Minimal Diagnosis            & n              & No             \\
		Nomystery                    & a,b,c,f                 & No             \\
		Permutation Pattern Matching & c,b                 & No             \\
		Ricochet Robots              & n,a,b,c              & No             \\
		Solitaire                    & a,b,c              & No             \\
		Stable Marriage              & -                & \textbf{Yes}   \\
		Valves Location              & n,a,c,f           & No             \\
		Weighted-Sequence            & n,c,b                & \textbf{Yes}   \\
		\aspccg~\enco;~\encm;~\encl                & n,a,b,c,f              & N/A   \\
		\hline
	\end{tabular}
	\label{tab:prog-details}
\end{table}

All tests were conducted on Ubuntu 18.04.3 with an Intel\textregistered~Xeon\textregistered~CPU E5-1620 v3 @ 3.50GHz and 32 GB of RAM. Furthermore, Python version 3.7.3 and \pyclingo version 5.4.0 are used to run \prd. 
Grounding and solving was  done by \clingo version 5.4.0. For all benchmarks execution was limited to 5 minutes.

\subsubsection{Intrinsic Evaluation}

\begin{table}[h]
	\centering
	\caption{Average error factor for benchmark programs, with and without keys}
	\begin{tabular}{lll}
		\hline
		\textbf{Program} & \textbf{Average Error Factor} & \textbf{Average Error Factor (Keyless)}   \\ \hline
		Hanoi Tower     & - & $1.5$  \\
		Nomystery     & $1.5$ & $1.5$ \\
		Permutation Pattern Matching$*$     & $\mathbf{3.8}$ & $5.0$ \\
		Solitaire     & -& $4.3$  \\
		Stable Marriage     & $\mathbf{3.7}$ & $7.5 * 10^5$ \\
		\hline
		Bottle Filling     & - & $4.9 * 10^9$  \\
		Incremental Scheduling$*$     & $1.1 * 10^5$ & $1.1 * 10^5$    \\
		Labyrinth$*$     & -& $1.3 * 10^1$  \\
		Minimal Diagnosis     & $8.2 * 10^3$ & $8.2 * 10^3$ \\
		Valves Location$*$     & $\mathbf{1.3 * 10^1}$ & $1.6 * 10^1$ \\
		\aspccg~\enco     & $\mathbf{2.9 * 10^1}$ & $3.1 * 10^1$ \\
		\aspccg~\encm     & $\mathbf{1.3 * 10^1}$ & $1.4 * 10^1$ \\
		\aspccg~\encl     & $2.2 * 10^1$ & $2.2 * 10^1$ \\
		\hline
		Knight Tour with Holes    & - & $1.9*10^{-4}$  \\
		Ricochet Robots     & ${2.0 * 10^{-1}}$ & $\mathbf{2.2 * 10^{-1}}$ \\
		Weighted Sequence     & ${6.0 * 10^{-3}}$ & $\mathbf{1.1 * 10^{-2}}$ \\
		\hline
	\end{tabular}
	\label{tab:acc-all}
\end{table}

Let $S$ be the true grounding size of an instance of a program computed by \gringo -- i.e., the number of rules in a ground program produced by \gringo.
Let $S'$ be the grounding size predicted by \prd for the same instance. We define a notion of an  \emph{error factor} on a program instance as  $S' / S$. The \emph{average error factor} of a program/benchmark is the average of all error factors across the instances of a program.
Table~\ref{tab:acc-all} shows the average error factor using \prdpr for all programs~\footnote{The numbers presented for the {\aspccg} \enco, \encm, \encl  are due to the original encoding of these benchmarks non-compatible with the ASP-Core-2 standard and utilized in the experiments by~\cite{lie21a}.}. The third column presents the case for programs when no key information is provided. Sign "-" indicates that for this benchmark no key information was provided within the main encoding.
The average error factor shown was rounded to make comparisons easier. An asterisk ($*$) next to a benchmark name indicates that not all 20 instances of this benchmark were grounded within the allotted time limit. For instance, 19  instances of the \emph{Incremental Scheduling}  benchmark
were successfully grounded, while the remaining instance timed out. For the  benchmarks annotated by $*$  we only report the average error factor assuming the instances grounded successfully. 

We partition the results into three groups using the average error factor. The partition is indicated by the horizontal lines on Table~\ref{tab:acc-all}. First, there are five programs where the estimates computed by \prd are, on average, less than one order of magnitude over. Second, there are eight programs that are, on average, greater than one order of magnitude over. Finally, three programs are predicted to have lower grounding sizes than in reality.

We also note the impact that keys have on certain programs. We especially emphasize the difference in error between \emph{Stable Marriage} with and without keys, where the average error factor is different by 5 orders of magnitude.
The numbers in bold mark instances in which information on keys change the prediction.

It is obvious that the accuracy of system \prd could still use improvements. In many cases the accuracy is drastically erroneous. These results are not necessarily surprising. We identify five main reasons for observed data on \prd:
\begin{enumerate}
    \item  Insufficient data modeling is one weak point of \prd. Since we do not keep track what actual constants could be present in the ground extensions of a predicate, it is often the case that we overestimate argument size due to our inability to identify repetitive values.
\item Since we only identified keys for root predicate arguments, many keys were likely missed; automatic key detection is the direction of future work.
\item System \prd has limited support for such common language extensions as aggregates. 
\item System \prd is vulnerable to what is known as \emph{error propagation} \citeb{ioannidis1991propagation}. 
\item	While one might typically expect \prd to overestimate due to its limited capabilities in detecting repeated data, the underestimation on \emph{Knight Tour with Holes},  \emph{Ricochet Robots}, and \emph{Weighted Sequence} programs is not surprising due to the fact that these programs are non-tight and utilize binary operations in terms.
\end{enumerate}

\subsubsection{Extrinsic Evaluation}
Here, we examine the \emph{relative} accuracy of system \prd alongside \pr and \lpopt. In other words, we measure the quality of \prd by analyzing the impact it has on \pr and \lpopt performance. We recall that in all experiments we consider that \prd is provided information on keys as documented  in Table~\ref{tab:keys}. 

Let $S$ be the grounding size of an instance of a program, where grounding is produced by \gringo. Let $S'$ be the grounding size of the same instance in a modified (rewritten) version of the program. In this context, the modified version will either be the logic program outputted after using \pr/\lpopt or the logic program outputted after using \prdpr/\prdlpopt. The \emph{grounding size factor} of a program's instance is defined as $S' / S$. As such, a grounding size factor greater than~$1$ indicates that the modification increased the grounding size, whereas a value  less than~$1$  indicates that the modification improved/decreased the grounding size. The \emph{average grounding size factor} of a benchmark is the average of all grounding size factors across the instances of a benchmark. While we target improving the grounding size of a program, the ultimate goal is to improve the overall performance of ASP grounding/solving. Thus, we also compare the execution time of the programs, as that is ultimately what we want to reduce.
Let~$S$ be the execution time of an answer set solver \clingo (including grounding and solving) on an instance of a benchmark. Let $S'$ be the execution time of \clingo on the same instance in a modified version of the benchmark. The \emph{execution time factor} of a program’s instance is defined as $S'/S$. The \emph{average execution time factor} of a benchmark is the average of all \emph{execution time factors} across the instances of a benchmark.

Table~\ref{tab:prprd-all} displays the average grounding size factor together with the {average execution time factor}
for \pr  and \prdpr  on all benchmark programs. An asterisk ($*$) following a program name indicates that not all 20 instances were grounded. In these cases, the average grounding size factor was only computed from instances where all 3 versions of the program (original, \pr, \prdpr) completed solving. 
The same concerns the computation of the average execution time factor. 
While we only consider instances in where all 3 version of the program completed grounding and then solving, we have included the exact number of instances grounded and solved by each version of the program, to show that the factors presented may be misleading. For example, consider program {\em Inc. Scheduling}, while \prdpr seems to have a slightly slower execution time than \pr alone, \prdpr managed to solve an additional instance, reflected by the decreased grounding time, therefore it would not be accurate to say the \pr outperformed \prdpr on that encoding.  A dagger ($\dagger$) following a program name indicates that there was a slight improvement for \prdpr, however this information was lost for the precision shown. 

\begin{table}[h]
	\centering
	\hspace{1mm}
	\begin{tabular}{l|ll|ll|lll}
		\multicolumn{1}{c}{}&
		\multicolumn{2}{c}{$\overbrace{\rule{7em}{0pt}}^{\text{Grounding Size Factor}}$}&
		\multicolumn{2}{c}{$\overbrace{\rule{7em}{0pt}}^{\text{Execution Time Factor}}$}\\
		\textbf{Program} & \textbf{\prshort}    & \textbf{\prdprshort} & \textbf{\prshort}    & \textbf{\prdprshort} & \textbf{Svd.} &  \textbf{Svd. \prshort} & \textbf{Svd. \prdprshort}    \\ \hline
		Hanoi Tower  & $1.41$ & $1.00$ & $1.67$ & $1.00$ & $20$ & $20$ & $20$    \\
		Inc. Scheduling$*$   & $1.14$ & $1.12$ & $1.06$ & $1.10$ & $13$ & $13$ & $14$   \\
		Minimal Diagnosis  & $1.06$ & $1.00$ & $1.04$ & $1.00$  & $20$ & $20$ & $20$  \\
		Solitaire    & $1.41$ & $1.00$ & $1.32$ & $0.99$ & $19$ & $19$ & $19$    \\
		Stable Marriage    & $0.13$ & $0.12$ &  $0.18$ & $0.17$ & $19$ & $19$ & $19$    \\
		\hline
		Bottle Filling     & $1.36$ & $1.36$ & $1.44$ & $1.43$  & $20$ & $20$ & $20$  \\
		Labyrinth    & $1.11$ & $1.11$ & $5.26$ & $5.27$  & $18$ & $18$ & $18$   \\
		Perm. Pattern Match.$*$ $\dagger$ & $0.13$ & $0.13$ & $0.14$ & $0.14$ & $16$ & $20$ & $20$\\
		Valves Location$\dagger$ & $1.00$ & $1.00$ & $1.03$ & $0.93$ & $3$ & $3$ & $3$  \\
		Weighted Sequence$\dagger$ & $1.00$ & $1.00$ & $3.05$ & $1.59$ & $19$ & $16$ & $17$    \\
		\aspccg~\enco     & $1.01$ & $1.01$ & $1.65$ & $2.28$ & $60$ & $60$ & $60$\\
		\hline
		\aspccg~\encm     & $0.90$ & $1.00$ & $1.57$ & $2.20$ & $60$ & $60$ & $60$\\
		\aspccg~\encl     & $0.70$ & $0.81$ & $1.71$ & $2.59$ & $60$ & $60$ & $60$\\
		Knight Tour with Holes    & $0.80$ & $0.90$ & $0.50$ & $2.45$ & $1$ & $1$ & $1$   \\
		Nomystery   & $0.62$ & $1.00$  & $1.23$ & $1.00$ & $7$ & $8$ & $7$    \\
		Ricochet Robots   & $0.91$ & $1.00$ & $0.85$ & $1.00$ & $20$ & $20$ & $20$   \\
		\hline
	\end{tabular}
	\caption{Average grounding size factors, and execution time factors for \prshort and \prdprshort}
	\label{tab:prprd-all}
\end{table}

\begin{table}[h]
	\centering
	\hspace{1mm}
	\begin{tabular}{l|ll|ll|lll}
	    \multicolumn{1}{c}{}&
		\multicolumn{2}{c}{$\overbrace{\rule{7em}{0pt}}^{\text{Grounding Size Factor}}$}&
		\multicolumn{2}{c}{$\overbrace{\rule{7em}{0pt}}^{\text{Execution Time Factor}}$}\\
		\textbf{Program} & \textbf{\lpopt} & \textbf{\prdlpopt} & \textbf{\lpopt} & \textbf{\prdlpopt} & \textbf{Svd.} & \textbf{Svd. \lpopt} & \textbf{Svd. \prdlpopt}    
		\\ \hline
		\aspccg~\enco     & $0.92$ & $0.89$ & $0.88$ & $0.87$ & $60$ & $60$ & $60$\\
		\aspccg~\encm     & $0.80$ & $0.75$ & $0.83$ & $0.79$ & $60$ & $60$ & $60$\\
		Hanoi Tower    & $1.41$ & $1.00$ & $1.59$ & $0.99$ & $20$ & $19$ & $20$      \\
		Minimal Diagnosis & $1.17$ & $1.00$ & $1.13$ & $1.00$ & $20$ & $20$ & $20$ \\   
		Bottle Filling & $1.00$ & $0.28$ & $0.98$ & $0.39$ & $20$ & $20$ & $20$  \\
		Valves Location & $1.00$ & $1.00$ & $1.00$ & $0.96$ & $3$ & $3$ & $3$    \\
		Solitaire$*$     & $1.03$ & $1.01$ & $4.53$ & $0.94$ & $18$ & $18$ & $18$     \\
		Knight Tour with Holes  & $3.36$ & $2.18$ & $1.28$ & $0.90$ & $1$ & $1$ & $1$  \\
		Labyrinth     & $1.24$ & $1.12$ & $10.45$ & $9.36$ & $18$ & $18$ & $18$    \\
		Weighted Sequence$\dagger$     & $1.07$ & $1.04$ & $1.13$ & $2.11$ & $19$ & $20$ & $20$    \\
		\hline
		Stable Marriage$\dagger$    & $1.01$ & $1.01$ & $1.02$ & $1.02$ & $19$ & $19$ & $19$    \\
		Perm. Pattern Match.$*$$\dagger$ & $0.14$ & $0.14$ & $1.15$ & $0.89$ & $16$ & $19$ & $20$    \\
		\hline
		Inc. Scheduling     & $1.78$ & $2.30$ & $1.01$ & $1.24$ & $13$ & $13$ & $13$    \\
		\aspccg~\encl     & $0.78$ & $0.87$ & $0.92$ & $0.93$ & $60$ & $60$ & $60$\\
		Nomystery    & $0.70$ & $0.95$  & $1.06$ & $2.72$ & $7$ & $8$ & $8$    \\
		Ricochet Robots     & $1.09$ & $1.18$ & $1.01$ & $2.02$ & $20$ & $20$ & $20$   \\
		\hline
	\end{tabular}
	\caption{Average grounding size factors, and execution time factors for \lpopt and \prdlpopt}
	\label{tab:prdlpopt-all}
\end{table}

\

We partition the results into three sets, indicated by the horizontal lines on Table~\ref{tab:prprd-all}. The first set denotes programs in which \prd improved the grounding size factor of the program, the second set denotes programs in \prd did not have a noticeable effect on the grounding size factor, and the last set denotes programs in which \prd harmed the grounding size factor of the program as compared to the rewriting without predictions. We note that there are five programs in which \prdpr reduces the grounding size when compared to \pr, five programs in which \prdpr does not impact the grounding size, and six
programs in which \prdpr increases the grounding size.
By grey highlight we mark the benchmarks where decrease in grounding size by means of using \prd resulted in the increase of solving time.

Table~\ref{tab:prdlpopt-all} displays the average grounding size factor together with the {average execution time factor}
for \lpopt  and \prdlpopt  on all benchmark programs. It is data is organized in the same style as within Table~\ref{tab:prprd-all} comparing
\pr  and \prdpr.
We note that there are ten programs in which \prdlpopt reduces the grounding size  when compared to \lpopt, two programs in which \prdlpopt does not impact the grounding size, and four programs in which \prdlpopt increases the grounding size.

Overall, the results illustrate the validity of \prd approach.
The system has especially positive impact within its integration with \lpopt. Also, the presented experimental data illustrates once more the importance of the development rewriting techniques and the possibility of their positive impact. Together with that decision support systems exemplified by \prd have to be designed and engineered to achieve the whole potential of ASP. We trust that system \prd is a solid step in that direction providing room for numerous improvements to account for nontrivial language features of ASP dialects. 


\section{Conclusions and Future Work}
We introduced a method for predicting grounding size of answer set programs. 
We implement the described method in stand-alone system \prd that runs agnostic to any answer set grounder/solver pair. We expect this tool to become a foundation to decision support systems for rewriting/preprocessing tools in ASP. Indeed, using \prd as a decision support guide to rewriting system \pr and \lpopt improves their outcome  overall. The same is observed for the case of the rewriting system called \lpopt.
This proves the validity of the proposed approach, especially as further methods for improving estimation accuracy are explored in the future. As such system~\prd is a unique tool unparalleled in earlier research ready for use within preprocessing frameworks in ASP.
As discussed in the introduction: this work provides an important step towards achieving a goal of {\em truly} declarative answer set programming. 

 The section on intrinsic evaluation indicated a number of potential areas worth of improving estimations. It is one of the future work directions. Another one is utilizing \prd within other preprocessing tools of ASP. We trust that both efforts can be now undertaken as a community effort given the availability and transparency of \prd. 
 Also, rather sophisticated techniques such as database-inspired
optimizations, back-jumping, rewritings, binder splitting techniques are available in modern implementations of grounders~\citeb{geb07b,cal17}. As of now these techniques are not accounted for when estimates are produced. Also at the moment,  uniform distribution of values between the maximum and minimum in predicate arguments is assumed. Looking into different assumptions is also an interesting future direction.

\paragraph{Acknowledgments}
We would like to thank    Mirek Truszczynski, Daniel Houston, Liu Liu, Michael Dingess, Roland Kaminski, Abhishek Parakh, Victor Winter, Parvathi Chundi, and Jorge Fandinno for valuable discussions on the subject of this paper.
 The work was partially supported by NSF grant 1707371. 
\paragraph{Competing interests:} The author(s) declare none.

%
%
%
\bibliographystyle{tlplike}

\bibliography{Bibliography-File}
\end{document}